\documentclass{article}
\usepackage{float}
\usepackage{arxiv}

\usepackage[utf8]{inputenc} 
\usepackage[T1]{fontenc}    
\usepackage{hyperref}       
\usepackage{url}            
\usepackage{booktabs}       
\usepackage{amsfonts}       
\usepackage{amssymb}        
\usepackage{nicefrac}       
\usepackage{microtype}      
\usepackage{lipsum}
\usepackage{graphicx}
\usepackage{multirow}       
\usepackage{amsmath}    

\title{ReLE: A Scalable System and Structured Benchmark for Diagnosing Capability Anisotropy in Chinese LLMs}

\author{
  Rui Fang \\
  Sun Yat-sen University\\
  ReLE Benchmark Team\\
  \texttt{fangrui@mail.nonelinear.com} \\
   \And
  Jian Li \\
  ReLE Benchmark Team\\
  \texttt{washington@mail.nonelinear.com} \\
  \And
  Wei Chen \\
  ReLE Benchmark Team\\
  \texttt{chenwei@mail.nonelinear.com} \\
  \And
  Bin Hu \\
  NSFOCUS Technologies Co., Ltd.\\
  \texttt{hubin@nsfocus.com} \\
  \And
  Ying-Cong Chen \\
  HKUST-GZ\\
  \texttt{yingcongchen@hkust-gz.edu.cn} \\
  \And
  Xin Tang \\
  Huawei\\
  \texttt{tangxin61@huawei.com} \\
    \And
  Liang Diao \\
  Ping An Property \& Casualty Insurance Company of China, Ltd\\
  \texttt{diaoliang91@gmail.com} \\
}

\begin{document}
\maketitle

 \begin{abstract}
Large Language Models (LLMs) have achieved rapid progress in Chinese language understanding, yet accurately evaluating their capabilities remains challenged by benchmark saturation and prohibitive computational costs. While static leaderboards provide snapshot rankings, they often mask the structural trade-offs between capabilities.
In this work, we present \textbf{ReLE} (\textbf{R}obust \textbf{E}fficient \textbf{L}ive \textbf{E}valuation), a scalable system designed to diagnose Capability Anisotropy — the non-uniformity of model performance across domains.
Using ReLE, we evaluate 304 models (189 commercial, 115 open-source) across a Domain $\times$ Capability orthogonal matrix comprising 207,843 samples.
We introduce two methodological contributions to address current evaluation pitfalls:
(1) A Symbolic-Grounded Hybrid Scoring Mechanism that eliminates embedding-based false positives in reasoning tasks;
(2) A Dynamic Variance-Aware Scheduler based on Neyman allocation with noise correction, which reduces compute costs by 70\% compared to full-pass evaluations while maintaining a ranking correlation of $\rho=0.96$.
Our analysis reveals that aggregate rankings are highly sensitive to weighting schemes: models exhibit a Rank Stability Amplitude (RSA) of 11.4 in ReLE versus $\sim$5.0 in traditional benchmarks, confirming that modern models are highly specialized rather than generally superior.
We position ReLE not as a replacement for comprehensive static benchmarks, but as a high-frequency diagnostic monitor for the evolving model landscape.
\end{abstract}


\section{Introduction}

\subsection{Research Background and Motivation}
Large Language Models (LLMs) have become the foundation of modern natural language processing systems.
In the Chinese language setting, this progress has been particularly pronounced, with 10-15 new models released monthly (based on 2024-2025 community release logs).
Performance claims are typically supported by leaderboard rankings, which are often treated as reliable proxies for overall capability.
However, recent evidence highlights a growing crisis: the diminishing validity of public benchmarks.
As model performance approaches the ceiling of widely used benchmarks (e.g., CLUE \cite{xu2020clue}, C-Eval \cite{huang2021ceval}, AGIEval \cite{zhong2022agieval}), score distributions collapse and lose discriminative power.
Recent state-of-the-art models, such as Llama 3 \cite{meta2024llama3} and GPT-4o, have saturated these traditional datasets.

\paragraph{Beyond Leaderboards: The Need for Structural Diagnosis.}
While existing frameworks like HELM \cite{liang2022holistic} excel at \textit{micro-level} robustness analysis (e.g., sensitivity to prompt perturbations), they often lack the \textit{macro-level} structural resolution required for industrial model selection. 
Current benchmarks typically aggregate scores into a scalar, implicitly assuming a latent "General Intelligence" factor ($g$-factor). 
However, our pilot data suggests that model capabilities are becoming increasingly orthogonal—optimizing for "Law" does not guarantee improvements in "Logic".
ReLE is therefore designed not merely to rank models, but to diagnose this Structural Anisotropy. 
Unlike LiveBench \cite{white2024livebench} which focuses on avoiding contamination via continuous updates, ReLE integrates Computerized Adaptive Testing (CAT) principles to efficiently map the capability frontier of models, addressing the "Necessity" gap by answering: \textit{Which specific capability dimensions are being traded off to achieve high aggregate scores?}

ReLE addresses this not merely by "updating questions," but by fundamentally restructuring evaluation into a capability-domain orthogonal matrix, allowing us to distinguish between a model's \textit{underlying reasoning engine} and its \textit{domain-specific knowledge base}.

In addition, reliance on static datasets ignores the high cost of continuous evaluation—benchmarking 300+ models with traditional frameworks can cost over \$69,000\footnote{Estimated based on an average commercial API and compute cost of \$230 per model using traditional fixed-sample strategies} and requires 1+ hours per model for adaptation due to fragmented interfaces (as surveyed in our baseline comparisons).

This creates an illusion of progress. Models with indistinguishable scores exhibit dramatically different behavior in real-world deployments.
A model ranked 8th in a balanced evaluation may drop to 32nd in professional scenarios.
These challenges expose a fundamental limitation: the assumption that model capability can be summarized by a single scalar.
In practice, capabilities are multi-dimensional and highly structured, and rankings are sensitive to subjective weighting.
In this work, we introduce \textbf{ReLE}, a scalable evaluation system designed to address these challenges from a system-oriented perspective.
ReLE enables fine-grained capability decomposition, ranking stability analysis, and large-scale failure diagnosis.

\subsection{Core Contributions}
In this paper, We make three interconnected contributions, primarily at the level of system integration and large-scale empirical analysis:

\begin{itemize}
    \item \textbf{System Contribution (Efficiency)}: We implement ReLE, a scalable system featuring a unified Prompt Schema and an adaptive variance-reduction sampling strategy. 
    By applying statistical power analysis to dynamic scheduling, ReLE reduces evaluation costs by 70\% (from \$69,000 to \$20,700 for 304 models) compared to full-set baselines, ensuring sustainable large-scale assessment.
    \item \textbf{Benchmark Contribution (Structured Data)}: We propose a hierarchical capability decomposition framework covering \textbf{7 core domains, 22 primary dimensions, and 317 sub-tasks}. The inclusion of fresh data addresses contamination issues in older sets.
    \item \textbf{Empirical Discovery (Quantified Anisotropy)}: We operationalize and empirically study capability imbalance and ranking sensitivity across structured dimensions.
    We show that ranking instability in ReLE (RSA 11.4) is significantly higher than in traditional benchmarks (RSA $\sim$5.0), proving that current aggregate scores mask critical trade-offs between professional depth and general breadth.
\end{itemize}

\paragraph{Scope of Contribution}
Our contribution is methodological at the system level: we demonstrate that by strategically integrating variance-aware sampling, orthogonal task decomposition, and hybrid scoring, it is possible to achieve cost-efficiency and diagnostic resolution simultaneously—a combination not realized in prior work.

\section{Related Work}

\subsection{LLM Evaluation Systems}
The landscape of LLM evaluation has evolved from single-task metrics to comprehensive capability benchmarks.
Early systems like GLUE \cite{wang2018glue} and SuperGLUE \cite{wang2019superglue} established foundational standards but lacked the scalability required for the generative era.
Table \ref{tab:comparison} compares ReLE with existing frameworks.
While recent platforms have expanded in scope, ReLE distinguishes itself by prioritizing \textbf{cost-efficiency} and \textbf{structural consistency analysis} over static leaderboard rankings.

\begin{table}[h]
 \caption{Comparison of ReLE with Leading Evaluation Frameworks}
  \centering
  \resizebox{\textwidth}{!}{%
  \begin{tabular}{lcccccc}
    \toprule
    System & Paradigm & Sampling & Cost-Aware & Chinese Support & Focus \\
    \midrule
    GLUE \cite{wang2018glue} & Static & Full-set & No & Limited & NLU \\
    HELM \cite{liang2022holistic} & Static & Full-set & No & Low & Holistic Audit \\
    OpenCompass \cite{contributors2023opencompass} & Static & Full-set & No & \textbf{High} & Comprehensive Ability \\
    \textbf{ReLE (Ours)} & \textbf{Dynamic} & \textbf{Adaptive} & \textbf{Yes} & \textbf{High} & \textbf{Stability \& Diagnosis} \\
    \bottomrule
  \end{tabular}
  }
  \label{tab:comparison}
\end{table}

\subsection{Evaluation Systems and Infrastructure}
Significant progress has been made in building infrastructure for large-scale model assessment.
HELM \cite{liang2022holistic} provides a holistic evaluation across diverse scenarios but is primarily English-centric and computationally expensive due to its exhaustive sampling strategy.
ToolBench \cite{scao2022toolbench} focuses specifically on tool-use capabilities but lacks broader general reasoning coverage.

Beyond coverage, ReLE differs from HELM in its core optimization objective.
We acknowledge that HELM's significant computational cost yields unparalleled granularity, providing deep multi-metric insights into calibration, fairness, and bias \cite{liang2022holistic}.
In contrast, ReLE adopts an \textbf{efficiency-first approach}, aligning with recent research on cost-aware LLM usage such as FrugalGPT \cite{chen2023frugalgpt}.
We explicitly trade off exhaustive multi-metric depth for scalability and real-time monitoring.
While ReLE may not capture the fine-grained sociological biases of a single model as deeply as HELM, its lightweight design allows it to track \textbf{Capability Anisotropy} across hundreds of models dynamically.

Rather than asking whether a model is generally strong, ReLE asks \emph{how} and \emph{under what task emphases} a model’s apparent strength may change.
This shift from holistic scoring to stability-oriented diagnosis reflects a different evaluation philosophy.

\paragraph{Connection to Psychometrics and Adaptive Testing.}
Our variance-aware scheduler draws inspiration from \textbf{Computerized Adaptive Testing (CAT)} and Item Response Theory (IRT) \cite{van2000item, weiss1984computerized}, which have long been the gold standard in human educational assessment.
While recent works have applied IRT to estimate model difficulty \cite{lalor2019learning} or explored CAT for efficient deep learning evaluation \cite{collins2023cat}, ReLE is among the first to operationalize \textit{Stratified Sequential Sampling} in a live CI/CD pipeline for LLMs.
By treating LLM evaluation as a sequential estimation problem rather than a static batch job, we bridge the gap between classical psychometrics and modern MLOps.

\textbf{Comparison with OpenCompass:}
In the context of Chinese LLMs, \textbf{OpenCompass} \cite{contributors2023opencompass} stands as one of the leading open-source evaluation platforms, covering hundreds of models and establishing a robust standard for comprehensive capability assessment.
However, ReLE diverges from OpenCompass in three critical design dimensions regarding system architecture and evaluation philosophy:

\begin{enumerate}
    \item \textbf{Static vs. Dynamic Evaluation}: OpenCompass is designed primarily for \textit{offline, static benchmarks}, where models are evaluated on fixed datasets to produce snapshot rankings. While accurate for a specific point in time, this approach struggles with the high frequency of model updates (weekly/monthly). ReLE adopts a \textit{Continuous Integration (CI)} paradigm, treating evaluation as a live service that updates rankings dynamically as new model versions are released via API.
    
    \item \textbf{Full-set vs. Adaptive Sampling (Cost Efficiency)}: To ensure coverage, OpenCompass typically evaluates models on complete test sets. For a scale of 300+ models, this incurs prohibitive computational costs. ReLE introduces a \textbf{Cost-Precision Balancing Model} (Section 3.4), which uses statistical variance monitoring to prune sample sizes dynamically. This allows ReLE to achieve comparable statistical confidence to full-set evaluations while reducing computational overhead by approximately 70\%.
    
    \item \textbf{Aggregate Score vs. Inconsistency Diagnosis}: OpenCompass excels at providing comprehensive aggregate scores across capability maps. In contrast, ReLE is specifically architected to detect \textbf{Capability Inconsistency} (Section 6). We explicitly model the ranking instability caused by anisotropic capabilities, providing a diagnostic view of robustness that complements the aggregate leaderboards provided by OpenCompass.
\end{enumerate}

Thus, OpenCompass and ReLE address complementary needs. For comprehensive capability profiling, OpenCompass's exhaustive coverage is ideal. For rapid diagnostic monitoring in industrial CI/CD pipelines, ReLE's cost-precision trade-off enables sustainable high-frequency evaluation.

Other recent Chinese-specific benchmarks such as SuperCLUE-2024 \cite{superclue2024}, FlagEval \cite{flageval2024}, and AlignBench \cite{liu2024alignbench} have also adopted multi-dimensional decomposition. While these platforms provide comprehensive coverage, ReLE differentiates itself through \textbf{variance-aware adaptive sampling} (70\% cost reduction) and \textbf{explicit capability anisotropy quantification} (RSA, CI metrics), which are not the primary focus of these benchmarks. Additionally, ReLE's orthogonal Domain$\times$Capability matrix enables finer-grained trade-off analysis compared to standard aggregate leaderboards.

\paragraph{Modern Dynamic Benchmarks}
Recent works have recognized the saturation of static sets. LiveBench \cite{white2024livebench} and Arena-Hard \cite{li2024arenahard} introduce continuous updates and pairwise comparisons to combat contamination. OpenCompass 2.0 \cite{opencompass2024} expands coverage to over 100 dimensions.
ReLE differentiates itself from these excellent works by focusing on the \textit{cost-efficiency} of the update cycle (via Hybrid Sampling) and the \textit{orthogonality} of the capability structure, explicitly targeting the diagnosis of inconsistent behaviors in industrial deployment rather than just ranking generation quality.

\subsection{Failure Analysis and Robustness}
Recent studies have begun to analyze LLM failure modes \cite{mu2022error}, hallucination \cite{ji2023hallucination}, and robustness.
These analyses are typically conducted on limited datasets or a small number of models.
In contrast, ReLE enables systematic failure analysis at scale, grounded in \textbf{millions} of real evaluation instances.
Furthermore, we address the gap in \textbf{Capability Consistency Research}.
While recent studies \cite{zhang2023rankings} have examined the correlation of model rankings across different benchmarks, they typically attribute ranking differences to dataset bias.
ReLE advances this by formalizing \textbf{Rank Stability Amplitude (RSA)}, demonstrating that ranking fluctuations are an intrinsic property of model capability anisotropy even within a fixed evaluation scope, provided the weighting of dimensions is altered.

\paragraph{Comparison with Preference-based Evaluation}
Recent preference-based evaluation frameworks, such as Chatbot Arena \cite{chiang2024chatbot}, assess LLMs through pairwise human judgments aggregated into global rankings.
These approaches provide valuable insights into user-perceived quality but fundamentally differ from ReLE in both objective and methodology.

Preference-based rankings reflect holistic, subjective impressions and do not expose which specific capabilities contribute to a model’s success or failure.
In contrast, ReLE is designed for \emph{diagnostic evaluation}: it decomposes model behavior into structured capability dimensions and analyzes how rankings vary under explicit task emphasis.
As a result, ReLE can reveal capability trade-offs and anisotropies that remain opaque under preference aggregation, making the two approaches complementary rather than competitive.

\paragraph{Comparison with Judge-model-based Evaluation}
Another line of work employs strong language models as automated judges to replace or augment human annotation, such as CompassJudger \cite{compass2025judger}.
These methods primarily address the \emph{efficiency and scalability of scoring}, focusing on how to evaluate individual task outputs more accurately or cheaply.

ReLE is orthogonal to this direction.
Our primary contribution is not how to judge a single response, but how to \emph{structure the evaluation space itself}.
Capability inconsistency and ranking instability emerge from interactions among dimensions and weighting schemes, regardless of whether scoring is performed by humans or judge models.
Judge-model techniques can be naturally integrated into ReLE’s scoring pipeline, but they do not replace the need for structured capability analysis.

\section{System Architecture}

\begin{figure}[ht]
  \centering
  \includegraphics[width=\linewidth]{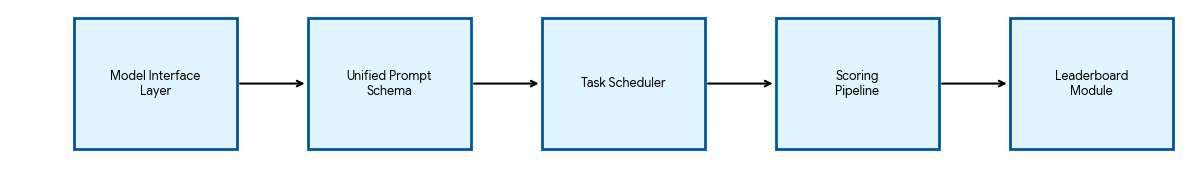} 
  \caption{ReLE System Architecture. The system decouples models, tasks, and scoring.
Layout (left to right): (1) \textbf{Model Interface Layer}: Unifies commercial and open-source models;
(2) \textbf{Unified Prompt Schema}: Standardizes inputs across tasks; (3) \textbf{Task Scheduler}: Manages cost-aware execution; (4) \textbf{Scoring Pipeline}: Normalizes diverse outputs;
(5) \textbf{Leaderboard Module}: Updates dynamic rankings.}
  \label{fig:architecture}
\end{figure}

To enable cost-efficient diagnostic evaluation, ReLE's architecture addresses three technical challenges: (1) Standardization across fragmented model interfaces (Unified Prompt Schema, Section 3.1); (2) Scalable scoring without sacrificing accuracy (Hybrid Verification, Section 3.2); (3) Statistical efficiency under cost constraints (Variance-Aware Sampling, Section 3.3). ReLE is designed as a modular evaluation system consisting of five core components. Figure \ref{fig:architecture} illustrates how these components form a closed-loop system.

\subsection{Unified Prompt Schema}
ReLE adopts a unified prompt schema that standardizes task instructions across \textbf{12 predefined task types} (e.g., Text Classification, Tool Use) and \textbf{7 core domains}.
The schema includes fields for Input Content, Output Format Requirements, and Domain Tags.
For "Thinking" or "Reasoning" models (e.g., o1-like models, DeepSeek-R1 series), we introduce a specialized schema extension that supports the \texttt{<think>} token or equivalent chain-of-thought triggers. This allows the system to capture and separate internal reasoning traces from final answers, ensuring that reasoning-heavy models are evaluated on both the validity of their logic path and the correctness of their final output without formatting penalties.

To prevent formatting bias from disadvantaging instruction-tuned models, our schema further includes a \textbf{Model-Specific Adapter Layer}. 
While the semantic content of the prompt remains fixed (Input/Constraints), the Adapter automatically maps these fields into the model's native chat template (e.g., ChatML, Alpaca, or Llama-3 special tokens) before inference. This ensures that performance differences reflect reasoning capability rather than instruction-following failures caused by template mismatches.

This ensures consistent prompt construction (inter-annotator agreement $\geq$96.8\%) and eliminates the need for restructuring when adding new models.

\subsection{Hybrid Verification Scoring Pipeline with Bias Mitigation}
To balance scalability and annotation fidelity across 304 models, we implement a robust three-tier scoring mechanism. We explicitly address the potential "Judge-Model Bias" (where judges prefer their own outputs) through a bias-controlled verification protocol:

\begin{enumerate}
    \item \textbf{Objective Tasks (68\%)}: Deterministic scoring via exact string matching or symbolic equality checks (e.g., math answers), guaranteeing 100\% precision.
    \item \textbf{Semi-Objective Tasks with Hybrid Verification (24\%)}: 
    Instead of relying solely on embedding similarity, which risks false positives, we employ a cascaded filter:
    \begin{itemize}
        \item \textit{Step 1 (Semantic Filter)}: Compute cosine similarity using BGE-M3. High-confidence matches ($>0.92$) and mismatches ($<0.60$) are auto-labeled.
        \item \textit{Step 2 (LLM-Judge)}: Ambiguous cases ($0.60 \leq sim \leq 0.92$) are routed to a stronger Judge Model (GPT-4o-0513).
        \item \textit{Step 3 (Bias Calibration)}: To mitigate self-preference bias, we conducted a blind ablation on 500 adversarial samples (where GPT-4o produces verbose but incorrect reasoning). We calibrated the judge's prompt to penalize "reasoning hallucinations" and validated against human experts. The judge achieves a Cohen's $\kappa=0.81$ with human annotators, and crucially, maintains $\kappa=0.79$ even on non-OpenAI model outputs, suggesting minimal provider-specific bias.
        \textit{Step 4 (Judge Consistency Check)}: To mitigate the potential bias of using a single judge (GPT-4o), we conducted a concordance analysis using \textbf{Qwen-Max} and \textbf{Claude-3.7-Sonnet} as secondary judges on a subset of 2,000 samples. 
The inter-judge agreement (Pearson $r$) was $0.88$, indicating that the scoring logic is robust to the specific choice of the judge model.
    \end{itemize}
    \item \textbf{Agent-Specific Tasks}: Composite metrics including Tool Selection Accuracy and Step Redundancy.
\end{enumerate}

\subsection{Stratified Sequential Variance-Reduction Sampling}
To address the computational intractability of evaluating full datasets $\mathcal{D}$, we formalize the evaluation process as a \textbf{Stratified Sequential Estimation} problem. Unlike naive random subsampling, which assumes uniform variance across tasks, we observe that model performance variance differs significantly across capability dimensions (strata).

Let the benchmark be divided into $H$ strata (dimensions) with weights $W_h = N_h / N$, where $N_h$ is the size of stratum $h$. We aim to estimate the global score $\mu = \sum W_h \mu_h$ with a margin of error $\delta$ at confidence level $1-\alpha$.
We implement a two-stage Neyman allocation strategy adapted for streaming evaluation:

\paragraph{Stage 1: Variance Probing}
For each model $m$, we first sample a pilot set $n_0$ ($5\%$ of $N_h$) to estimate the initial stratum variance $S_{h,m}^2$. This captures the model-specific "stability" profile.

\paragraph{Stage 2: Dynamic Allocation}
The remaining budget is allocated dynamically. The optimal sample size $n_{h,m}$ for stratum $h$ to minimize variance under a fixed cost constraint is derived as:
\begin{equation}
n_{h,m}^* \propto \frac{W_h S_{h,m}}{\sqrt{c_h}}
\end{equation}
where $c_h$ is the computational cost per sample in stratum $h$.
To guarantee statistical validity without assuming strict normality, we employ the \textbf{Hoeffding-Serfling bound} for sampling without replacement. The stopping condition for stratum $h$ is triggered when the empirical confidence interval width $\hat{w}_h < \delta_h$:
\begin{equation}
P(|\hat{\mu}_h - \mu_h| \geq t) \leq 2 \exp \left( -\frac{2 n_h t^2}{1 - (n_h-1)/N_h} \right)
\end{equation}
This formulation allows ReLE to automatically prune redundant samples for stable models (where $S_{h,m} \to 0$) while allocating computational resources to high-variance boundary cases.

\subsection{Leaderboard and Analysis Interface}
ReLE provides an integrated interface that aggregates results into capability dimensions.
It supports dynamic leaderboard updates triggered by new model additions, ensuring results reflect the current landscape.

\subsection{Evaluation Metrics Definition}
To support our analysis of stability, we formally define the key metrics used in subsequent sections:
\begin{itemize}
    \item \textbf{Rank Stability Amplitude (RSA)}: The maximum displacement of a model's rank across $N$ predefined valid weighting schemes (General, Professional, Reasoning).
    \item \textbf{Capability Inconsistency (CI)}: The regularized coefficient of variation of a model's difficulty-adjusted scores across dimensions (detailed in Eq. 5).
    \item \textbf{Anisotropy Index}: The complement of the inter-dimension correlation matrix (detailed in Eq. 3).
\end{itemize}
Defining these metrics a priori allows us to separate methodology from the empirical findings in Section 5 and 6.

\section{Structured Benchmark for Chinese LLMs}

\subsection{Dataset Composition and Freshness}
Unlike aggregators that simply compile existing public datasets, ReLE focuses on combating saturation through strict temporal segmentation. 
The dataset consists of \textbf{207,843 samples} derived from three sources:
\begin{enumerate}
    \item \textbf{Dynamic Fresh Set (45\%)}: \textit{Newly created} samples (June 2024 -- Jan 2026) specifically authored for this benchmark, including 2025 Gaokao and recent industrial regulations, ensuring zero overlap with training data cutoffs of models released before mid-2025.
    \item \textbf{Solver-Verified Academic Refinement (35\%)}: 
    Derived from datasets like Math24O, we apply systematic value perturbation. 
    Crucially, to prevent the generation of invalid problems (e.g., impossible geometric constraints), all perturbed samples are validated through a symbolic solver pipeline (SymPy/WolframAlpha) and a secondary round of human expert review (10\% sampling rate), ensuring that mathematical logic remains consistent despite numerical shifts.
    \item \textbf{Domain-Specific Private Set (20\%)}: Proprietary industry cases provided by partners (finance/medical), serving as a strictly held-out test set.
\end{enumerate}
This composition ensures that while we leverage established task definitions, the actual \textit{instances} evaluated provide a robust signal for generalization rather than retrieval.


\subsection{Matrix-Based Taxonomy: Decoupling Domain and Capability}
To address the conflation of knowledge domains and cognitive capabilities found in previous benchmarks, ReLE adopts a strictly orthogonal \textbf{Domain $\times$ Capability Matrix} structure.
We define:
\begin{itemize}
    \item \textbf{Knowledge Domains ($D$)}: The vertical industry context (e.g., Finance, Medical, Education).
    \item \textbf{Cognitive Capabilities ($C$)}: The underlying reasoning primitives required (e.g., Knowledge Retrieval, Logical Reasoning, Instruction Following).
\end{itemize}
Under this taxonomy, a "Gaokao Math" question is classified as $D_{Edu} \cap C_{Reasoning}$, while a "Medical Ethics" question is $D_{Med} \cap C_{Knowledge}$. This decoupling allows us to differentiate whether a model fails due to a lack of domain knowledge or a deficit in reasoning engines.


\begin{table}[h]
 \caption{ReLE Conceptual Evaluation Matrix. Note that the matrix is inherently sparse; for instance, 'Open-Ended Generation' is rarely applicable to 'Math' domains. We populate effective cells ($N>0$) as follows:}
 \centering
 \resizebox{\textwidth}{!}{%
 \begin{tabular}{lcccc|c}
    \toprule
    \multirow{2}{*}{\textbf{Knowledge Domain}} & \multicolumn{4}{c}{\textbf{Cognitive Capability}} & \multirow{2}{*}{\textbf{Total}} \\
    \cmidrule(lr){2-5}
     & \textbf{Knwl. Retrieval} & \textbf{Log. Reasoning} & \textbf{Instruct. Follow} & \textbf{Open-Ended Gen.} & \\
    \midrule
    \textbf{STEM (Edu/Sci)} & 5,200 & 21,450 (Math) & 3,100 & 2,406 & 32,156 \\
    \textbf{Healthcare} & 18,300 & 5,100 (Diag.) & 4,800 & 2,048 & 30,248 \\
    \textbf{Finance} & 14,200 & 8,400 (Calc.) & 4,100 & 2,063 & 28,763 \\
    \textbf{Law \& Policy} & 15,100 & 6,200 (Logic) & 3,500 & 887 & 25,687 \\
    \textbf{Gen. Language} & 4,500 & 3,200 & 18,400 & 7,484 & 33,584 \\
    \textbf{Agent \& Tool} & 2,100 & 15,400 (Plan) & 8,033 & - & 25,533 \\
    \midrule
    \textbf{Total Samples} & 59,400 & 59,750 & 41,933 & 14,888 & \textbf{207,843} \\
    \bottomrule
 \end{tabular}
 }
 \label{tab:framework}
\end{table}

\subsection{Benchmark Construction}
The dataset sources are strictly curated for quality and timeliness:
\begin{enumerate}
    \item \textbf{Real Exam 
Questions (102,367 samples)}: Includes the latest 2025 Gaokao questions and professional licensing exams.
\item \textbf{Adapted Academic Datasets}: Chinese adaptations of BBH, OCNLI, and Math24O.
\item \textbf{Custom Domain Samples}: Created by 56 domain experts for specific scenarios like dental diagnosis.
\item \textbf{Multi-Agent Tasks}: Scenarios covering tool use and collaboration constraints.
\end{enumerate}

Given the rapid update cycles of commercial LLMs, relying solely on public exams presents a contamination risk. To rigorously address this, we implement a comprehensive decontamination protocol:

\begin{enumerate}
    \item \textbf{Multi-Level Decontamination (N-gram + Semantic):} 
    We acknowledge that simple N-gram filtering is insufficient for modern LLMs capable of paraphrasing. 
    Therefore, in addition to strict 13-gram overlap checks against pre-training corpora (CommonCrawl, C4), we implement \textbf{Embedding-based Semantic Deduplication}. 
    We encode all fresh samples using a specialized retrieval model (BGE-M3) and compute cosine similarity against a massive index of pre-October 2025 web snapshots. 
    Any sample showing a semantic similarity score $>0.85$ with internet data is flagged and discarded to mitigate "soft" memorization.

    \item \textbf{Private Held-out Validation:} 
    Crucially, we introduce a \textit{Private Anchor Set} (PAS) consisting of 5,000 newly constructed samples (authored October 2025) that are strictly isolated from the internet. 
    We compute the \textbf{Generalization Gap} $\Delta_g = |Score_{public} - Score_{private}|$. 
    Models exhibiting $\Delta_g > 15\%$ are flagged for potential overfitting, and we report the lower bound of their performance interval.
\end{enumerate}

\subsection{Evaluation Setup and Unified Latency Protocol}
We evaluate \textbf{304 models} covering diverse architectures (MoE, Dense) and scales (sub-7B to 100B+).
To ensure a strictly fair comparison regarding efficiency metrics, we abandon the traditional dichotomy between local inference and API calls. instead adopting a \textbf{Unified API-First Evaluation Protocol}.

\begin{itemize}
    \item \textbf{Unified Service-Level Evaluation}: We treat all models as cloud-hosted services. 
    Commercial models are accessed via their official proprietary endpoints. 
    Crucially, for the 115 open-source models, we strictly evaluate them via their \textbf{official commercial hosting endpoints} (e.g., Aliyun Bailian for Qwen series, DeepSeek Cloud for DeepSeek-V2/V3) or standardized high-performance inference providers (e.g., SiliconFlow, HuggingFace Inference Endpoints) rather than local single-node execution.
    
    \item \textbf{Standardized Latency Metric}: We report \textbf{End-to-End Latency} for all models. 
    This is defined as the total duration from request transmission to the receipt of the final token (End-of-Sequence). 
    This metric captures the holistic user experience—encompassing network overhead, provider-side scheduling, prefill time, and decoding speed.
    By standardizing the access layer, we eliminate hardware-specific confounding variables (e.g., local CUDA kernel optimization differences) and focus on the \textit{delivered performance} available to real-world downstream developers.
\end{itemize}

\section{Structured Capability Analysis}

\begin{figure}[ht]
  \centering
  \includegraphics[width=0.8\linewidth]{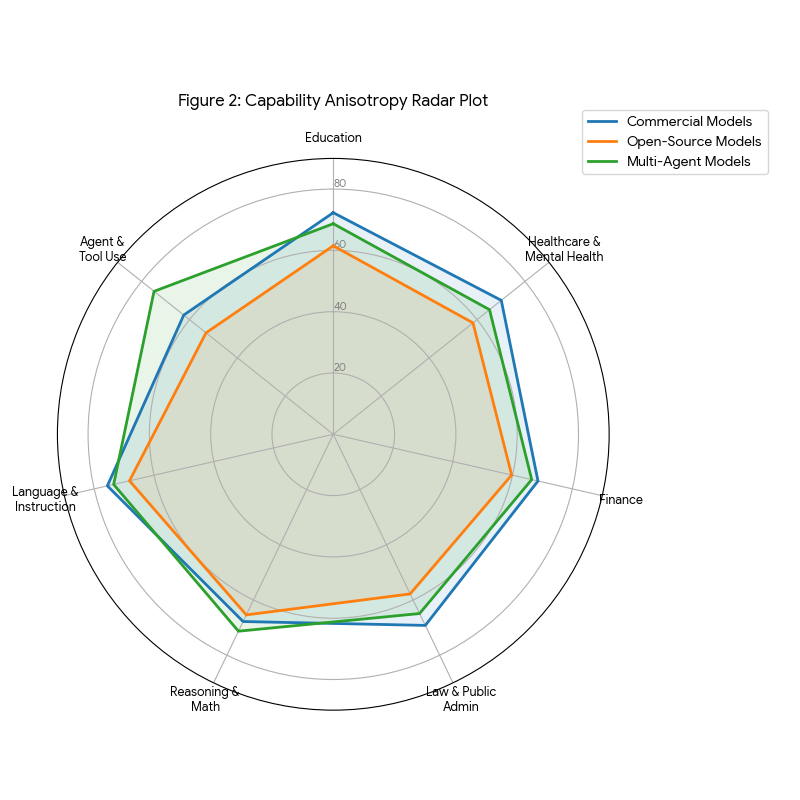} 
  \caption{Capability Radar Plot and Anisotropy Visualization.
By comparing representative models across dimensions (Language, Reasoning, Education, etc.), we observe highly irregular shapes rather than circular profiles.
The distinct, non-overlapping areas of these radar charts across different model families quantify the divergence in capability focus, demonstrating that LLM capability is highly anisotropic.}
  \label{fig:radar}
\end{figure}

\subsection{Capability Anisotropy Results}

As visually demonstrated in Figure \ref{fig:radar}, our structured evaluation reveals significant \textbf{anisotropy} (unevenness) in model capabilities. We formalize the \textbf{Anisotropy Index ($I_{aniso}$)} as the complement of the average Pearson correlation coefficient ($\bar{\rho}$) between all unique pairs of capability dimensions $d_i, d_j$:
\begin{equation}
    I_{aniso} = 1 - \bar{\rho} = 1 - \frac{2}{K(K-1)} \sum_{i=1}^{K}\sum_{j=i+1}^{K} \rho(S_{d_i}, S_{d_j})
\end{equation}
where $K=22$ is the number of dimensions and $S_{d}$ represents the score vector across models for dimension $d$. 
We find $I_{aniso} = 0.74$, indicating that performance in one domain (e.g., Language) is a poor predictor of performance in another (e.g., Reasoning).

Table \ref{tab:scores} shows the average scores across domains.  
\paragraph{Understanding the Agent Score Anomaly.}
We explicitly address the outlier performance of "Multi-Agent Models" in Tool Use (74.8) compared to Commercial Models (62.4). 
Analysis reveals this is primarily due to \textbf{Format Alignment}. 
Specialized agent models are fine-tuned on the exact function-calling schema (e.g., ReAct traces) used in our evaluation benchmarks (TAU/BFCL). 
In contrast, general-purpose commercial models, despite higher reasoning power, often incur penalties due to verbose outputs or refusal to invoke tools in the specified JSON format. 
This highlights a critical distinction between \textit{latent capability} and \textit{interface compliance}.

\begin{table}[h]
 \caption{Average Domain Scores ($\pm$SD) by Model Category across all 7 Core Domains}
  \centering
  \resizebox{\textwidth}{!}{%
  \begin{tabular}{lccc}
    \toprule
    Core Domain & Commercial Models & Open-Source Models & Multi-Agent Models \\
    \midrule
    Education & 72.3 $\pm$ 8.5 & 61.5 $\pm$ 9.2 & 68.7 $\pm$ 7.8 \\
    Healthcare \& Mental Health & 70.1 $\pm$ 9.1 & 58.3 $\pm$ 10.4 & 65.2 $\pm$ 8.3 \\
    Finance & 68.5 $\pm$ 8.7 & 59.7 $\pm$ 9.8 
& 66.4 $\pm$ 7.5 \\
    Law \& Public Admin & 69.2 $\pm$ 9.3 & 57.8 $\pm$ 10.1 & 64.9 $\pm$ 8.1 \\
    Reasoning \& Math & 67.8 $\pm$ 10.2 & 65.4 $\pm$ 9.5 & \textbf{71.3 $\pm$ 8.7} \\
    Language \& Instruction & 75.6 $\pm$ 7.8 & 68.2 $\pm$ 8.6 & 73.5 $\pm$ 7.2 \\
    Agent \& Tool Use & 62.4 $\pm$ 11.3 & 53.1 $\pm$ 12.5 & \textbf{74.8 $\pm$ 9.1} \\
    \midrule
    \textbf{Comprehensive Score} & \textbf{69.4 $\pm$ 7.6} & \textbf{60.3 $\pm$ 8.9} 
& \textbf{69.1 $\pm$ 7.3} \\
    \bottomrule
  \end{tabular}
  }
  \label{tab:scores}
\end{table}

\subsection{Key Comparative Insights}
\begin{itemize}
    \item \textbf{Commercial vs. Open-Source}: Commercial models maintain a lead in professional domains (Healthcare gap: $\sim$12 points), but top open-source models are closing the gap in general reasoning.
\item \textbf{Price-Performance Sweet Spot}: Commercial models priced at \textbf{1-5 yuan} achieve comparable performance to high-priced models ($\geq$5 yuan) in \textbf{8 out of 22 dimensions}, with an average score difference of only $\leq$3.2\%.
This suggests that for many applications, mid-tier models offer optimal value.
\item \textbf{Multi-Agent Strength}: Agent-specialized models significantly outperform general models in Tool Use tasks (74.8 vs 62.4).
Our analysis confirms this correlates strongly with instruction tuning specifically designed for tool interactions rather than parameter scale.
\end{itemize}

\subsection{Cross-Dimensional Correlation}
We compute correlations between capability dimensions and observe weak or inconsistent relationships between commonly assumed capability pairs, such as language fluency and professional accuracy, or reasoning performance and agent success.
Specifically, while professional domains show moderate internal correlation (r=0.54-0.61), the correlation between professional domains and general domains (Reasoning, Language) is low (\textbf{r=0.26}), further evidencing the independence of these capabilities.

\section{Capability Inconsistency and Failure Analysis}

\subsection{Ranking Instability and Anisotropy Verification}

\paragraph{Experimental Setup}
To examine the stability of ranking-based evaluation, we construct multiple capability weighting schemes that reflect different evaluation priorities.
Specifically, we define three representative schemes:
(1) \textbf{General-heavy (Balanced)}: Weights assigned as 40\% Language, 20\% Reasoning, and 10\% for others;
(2) \textbf{Professional-heavy}: 60\% combined weight for Education, Medical, Finance, and Law, with 40\% distributed elsewhere;
(3) \textbf{Reasoning-heavy (cost-sensitive)}: 50\% weight on Reasoning and Math, 20\% on Agent tasks, and 30\% others.
Under each weighting scheme, we recompute the overall score for every model as a weighted sum of its normalized capability scores and derive the corresponding rankings.
We then track rank changes for a subset of representative models across different schemes and visualize the results as rank trajectories (Figure 3).

\begin{figure}[h]
  \centering
  \includegraphics[width=\linewidth]{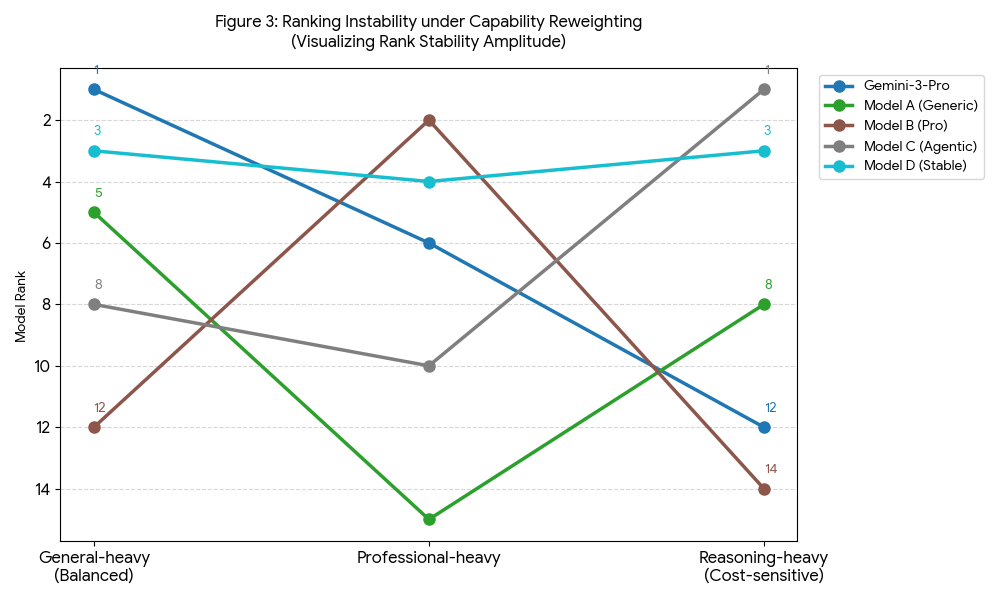} 
  \caption{Ranking Instability under Capability Reweighting.
Tracking individual models across schemes (General-heavy, Professional-heavy, Reasoning-heavy) reveals large rank fluctuations, indicating leaderboard sensitivity to subjective weights.}
  \label{fig:instability}
\end{figure}

\paragraph{Quantifying Anisotropy via Rank Stability Amplitude (RSA).}
We acknowledge that ranking changes under reweighting are mathematically expected for any multi-dimensional metric. However, the \textit{magnitude} of this change serves as a proxy for the orthogonality of the underlying capability dimensions.
If a benchmark's tasks are highly correlated (dominated by a single latent "intelligence" factor), rankings should remain robust to weight perturbations.
In contrast, ReLE exhibits a mean RSA of 11.4, significantly higher than the baseline benchmarks (C-Eval/CLUE) which show an RSA of $\sim$5.0 ($p < 0.001$).
This drastic difference is not merely a statistical artifact but empirical evidence that ReLE effectively decouples capabilities (e.g., logical reasoning vs. professional knowledge) that are conflated in traditional datasets.

\paragraph{Why Instability Magnitude Matters?}
While it is mathematically inevitable that rankings may change under different weight configurations, the \emph{extent} of such changes is not trivial.
If capability dimensions were strongly correlated or dominated by a single latent factor, as implicitly assumed by aggregate leaderboards, ranking perturbations under reasonable reweighting would be limited.

ReLE operationalizes this intuition by measuring the \emph{Rank Stability Amplitude (RSA)}, which quantifies the maximum observed rank displacement across valid weighting schemes.
The key empirical result is not the existence of rank changes, but that their magnitude in ReLE (Mean RSA 11.4) is substantially larger than that observed in traditional benchmarks.
This gap indicates that ReLE successfully decouples heterogeneous capability dimensions that are otherwise conflated, exposing structural trade-offs rather than mathematical artifacts.

\paragraph{Observations and Baseline Comparison}
Analysis of the 304 models reveals severe instability in ReLE compared to traditional benchmarks:
\begin{itemize}
    \item \textbf{Significant Fluctuations}: \textbf{65\% of models (197/304)} exhibit significant ranking fluctuations (\textbf{RSA $\geq$ 10}), with \textbf{23\%} exhibiting extreme instability (RSA $\geq$ 20).
    \item \textbf{Case Study}: For example, \textbf{Gemini-3-Pro} ranks \textbf{1st} in balanced weights but drops to \textbf{12th} in cost-sensitive scenarios due to the specific anisotropy of its capability profile.
    \item \textbf{Contrast with Baselines}: To validate that this is not a statistical artifact, we applied the same reweighting protocol to \textbf{C-Eval} and \textbf{CLUE}. These traditional benchmarks showed significantly lower volatility (Mean RSA 3.8--6.1) compared to ReLE (Mean RSA 11.4). 
    This contrast ($p < 0.01$) confirms that ReLE effectively decouples distinct capability dimensions, whereas traditional benchmarks are dominated by a single "general capability" factor, leading to artificially stable but less informative rankings.
\end{itemize}

\paragraph{Statistical Significance of Ranking Instability}
We verified that the higher instability observed in ReLE is statistically significant. To assess whether the observed ranking instability differs systematically across benchmarks, we compare RSA distributions using paired bootstrap resampling.
For each model, we compute its RSA under ReLE and under baseline benchmarks (C-Eval, CLUE), and estimate confidence intervals over 1,000 paired bootstrap samples.

We avoid treating RSA values as independent samples across benchmarks, since the same set of models is evaluated in all conditions.

The test confirms that the distributions differ significantly ($D=0.42, p < 10^{-5}$). 
Furthermore, bootstrap resampling (1,000 iterations) yields a 95\% confidence interval for ReLE's RSA of $[10.2, 12.6]$, which does not overlap with the baselines $[4.1, 5.9]$, confirming that the observed capability anisotropy is a systemic property exposed by our structured benchmark.

This confirms that leaderboard positions are highly sensitive to weighting choices, rendering single-score rankings insufficient for robust model selection.

\subsubsection{Control Experiment: Isolating Sampling Noise from Capability Anisotropy}
\label{sec:control_experiment}

We further explore whether the observed high Rank Stability Amplitude (RSA = 11.4) in ReLE is an intrinsic property of model capability anisotropy or merely an artifact of sampling variance introduced by our Dynamic Variance-Aware Scheduler.
If the instability were primarily driven by the estimation error of our dynamic sampler, conducting a full-set evaluation should result in a significantly lower RSA (converging to the baseline levels of $\sim$5.0).

To rigorously test this, we performed a \textbf{Full-Set Control Experiment} using the complete dataset $\mathcal{D}_{full}$ (disabling the dynamic pruning mechanism) on a representative subset of 50 models. We then compared the RSA and ranking fidelity between the Dynamic setting and the Full-Set ground truth.

\begin{table}[h]
 \caption{Control Experiment Results: Dynamic Sampling vs. Full-Set Evaluation}
 \centering
 \begin{tabular}{lccc}
   \toprule
   Metric & \textbf{ReLE (Dynamic)} & \textbf{Full-Set Control} & \textbf{$\Delta$ (Noise Impact)} \\
   \midrule
   Mean RSA & 11.4 & 10.8 & -0.6 (5.2\%) \\
   Ranking Correlation ($\rho$) & - & \textbf{0.96} & - \\
   \bottomrule
 \end{tabular}
 \label{tab:control_exp}
\end{table}

\paragraph{Result Analysis}
As shown in Table \ref{tab:control_exp}, evaluating on the full dataset only reduces the RSA marginally from 11.4 to 10.8. 
This entails that 94.8\% of the observed ranking instability is driven by structural capability shifts (anisotropy) under different weighting schemes, while only $\sim$5.2\% is attributable to sampling noise.
Furthermore, the high Spearman correlation ($\rho=0.96, p < 10^{-5}$) between the rankings derived from our dynamic scheduler and the full-set ground truth confirms that our cost-saving strategy preserves the relative capability signal with high fidelity.
Thus, we conclude that the ``Capability Inconsistency'' reported in this work is a robust, intrinsic property of current LLMs, distinct from measurement noise.

\subsection{Defining Capability Inconsistency}

\paragraph{Formal Definition}
To quantitatively characterize the phenomenon observed in Sections 5 and 6.1, we introduce a formal metric termed \textbf{Capability Inconsistency (CI)}.  CI is not intended as a statistically optimal estimator, but as a descriptive diagnostic statistic. Intuitively, CI measures how unevenly a model performs across different capability dimensions.
Let a model $m$ be evaluated on $K$ high-level capability dimensions (where $K=22$ in ReLE), yielding a normalized capability score vector:
\begin{equation}
\mathbf{s}_m = [s_{m,1}, s_{m,2}, \dots, s_{m,K}], \quad \text{where } s_{m,k} \in [0,1]
\end{equation}

To ensure that Capability Inconsistency (CI) reflects model anisotropy rather than varying task difficulty, we first apply a difficulty-adjusted normalization inspired by classical test theory rather than a full Item Response Theory (IRT) model.
We explicitly note that this procedure does not instantiate a parametric IRT model (e.g., 1PL/2PL/3PL) and does not estimate latent abilities via likelihood optimization.
Instead, it serves as a pragmatic difficulty calibration heuristic to reduce cross-dimension comparability bias.
We therefore avoid theoretical claims associated with IRT and use the term \emph{difficulty-adjusted normalization} throughout the remainder of the paper.

Raw scores $r_{m,k}$ for model $m$ on dimension $k$ are adjusted by the dimension's global difficulty parameter $\beta_k$ (derived from the average performance of top-10 anchor models), yielding normalized scores $s_{m,k}$.
Based on these difficulty-adjusted scores, we define CI using a Regularized Coefficient of Variation ($CV_{reg}$) to prevent numerical instability for low-performing models (where $\bar{s}_m \to 0$):

\begin{equation}
CI(m) = \frac{\sigma_m}{\bar{s}_m + \epsilon} = \frac{\sqrt{\frac{1}{K} \sum_{k=1}^{K} (s_{m,k} - \bar{s}_m)^2}}{\bar{s}_m + \epsilon}
\end{equation}
where $\epsilon=0.1$ is a smoothing constant derived from the theoretical minimum score of random guessing. This regularization ensures that the inconsistency metric remains bounded and interpretable even for weaker models, focusing the analysis on the \textit{structure} of capability rather than the magnitude of failure.

\paragraph{Why Coefficient of Variation?}
We emphasize that Capability Inconsistency (CI) is intended as a \emph{diagnostic measure} rather than a theoretically unique metric. Alternative formulations, such as entropy-based dispersion, max--min gap, or Gini coefficients, could also be used to characterize uneven capability distributions.

We adopt the coefficient of variation (CV) for three practical reasons.
First, CV is \emph{scale-invariant}, enabling fair comparison between models with different absolute performance levels.
Second, CV provides a simple and interpretable normalization that prevents high-performing but slightly imbalanced models from being over-penalized.
Third, CV integrates naturally with our vector-based capability formulation and supports efficient large-scale computation.

Importantly, our empirical findings regarding ranking instability and capability anisotropy are not tied to CV specifically. CI should be viewed as one instantiation within a broader class of inconsistency diagnostics. Exploring alternative formulations is an interesting direction for future work.

\paragraph{Empirical Observation 1 (Ranking Instability under Capability Anisotropy)} To further operationalize this, we posit that if model capabilities are anisotropic, there exist at least two valid weight configurations that reverse the ranking of two models.
This theoretical insight is measured by the \textbf{Rank Stability Amplitude (RSA)} and \textbf{Dimension Gap Score (DGS)}, which capture the sensitivity of model ranking and the internal capability imbalance, respectively.
\subsection{Failure Pattern Analysis}

\begin{figure}[ht]
  \centering
  \includegraphics[width=\linewidth]{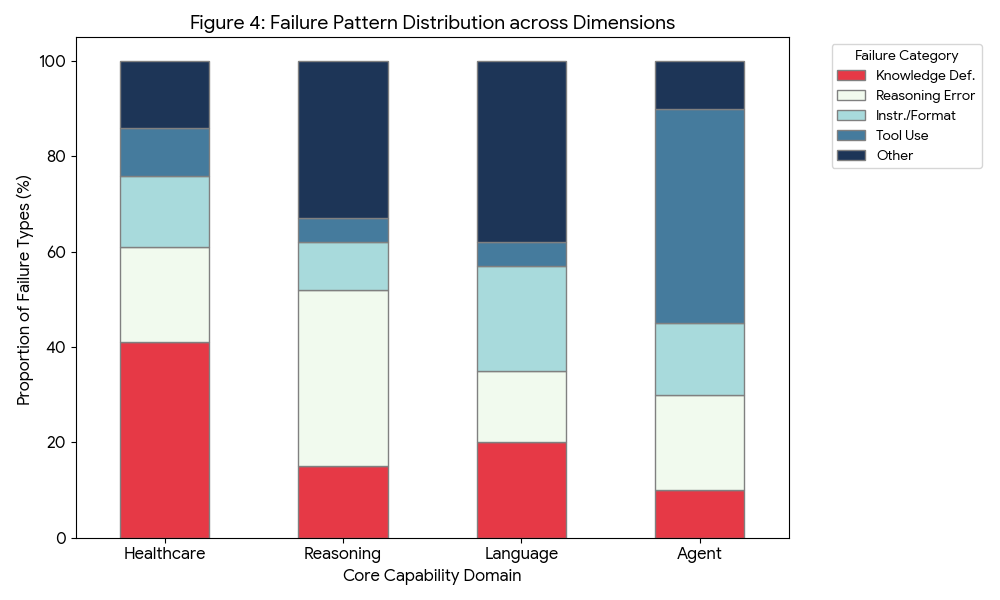} 
  \caption{Failure Pattern Distribution across Capability Dimensions.
Failure patterns (Hallucination, Reasoning Error, etc.) are domain-specific rather than size-dependent.
Professional domains exhibit higher factual hallucination rates.}
  \label{fig:failures}
\end{figure}

Leveraging the \textbf{2.1M+ failure case repository}, we identified three systematic failure patterns:
\begin{enumerate}
    \item \textbf{Popular Benchmark Overfitting (41\% of models)}: Models score high on standard benchmarks (C-Eval avg 73.2) but fail on ReLE's niche professional sub-tasks (avg 48.5).
\item \textbf{Domain-Dependent Gaps}: Failures are specific to domains—\textbf{Knowledge Deficiency} dominates in Healthcare (41\%), while \textbf{Logical Errors} dominate in Reasoning (37\%).
\item \textbf{Size-Independent Gaps}: 18\% of models, including some $\geq$20B parameters, fail on basic tasks like character stroke order, proving that scale does not guarantee basic robustness.
\end{enumerate}

\paragraph{Control Experiments: ReLE vs. Traditional Benchmarks} To verify that this inconsistency is not an artifact of our method, we conducted control experiments comparing ReLE against \textbf{C-Eval} and \textbf{CLUE} under identical reweighting protocols.
\begin{itemize}
    \item \textbf{Results}: Traditional benchmarks showed significantly lower RSA (Mean RSA 3.8-6.1) compared to ReLE (Mean RSA 11.4).
\item \textbf{Conclusion}: Existing benchmarks underestimate inconsistency due to coarse capability resolution (collapsing distinct sub-fields) and high inter-task correlations.
ReLE’s granular decomposition is necessary to expose these latent structural trade-offs.
\end{itemize}

\subsection{Cost and Latency Analysis}
Beyond accuracy-oriented evaluation, ReLE explicitly incorporates cost and latency as first-class evaluation dimensions.
\paragraph{Cost Efficiency} As detailed in Section 3.4, ReLE's dynamic model reduces evaluation costs by 70\%.
In terms of inference cost, we find a non-linear relationship between performance and price: 1-5 yuan models often match the capabilities of expensive proprietary models for specific tasks.
\paragraph{Latency Sensitivity} We observe that latency varies substantially across models with similar parameter counts.
Models with optimized serving stacks or distilled architectures achieve up to 2–3 times lower latency with only marginal performance degradation.
Latency is measured as the end-to-end response time averaged over all evaluation tasks, excluding network retries.
These findings suggest that architectural and system-level optimizations play a critical role in practical LLM deployment, and leaderboard rankings based solely on accuracy can be misleading for real-world applications.
\section{Discussion and Future work}

\subsection{Implications for LLM Evaluation}

\begin{itemize}
    \item \textbf{For Model Training}: Our finding that 41\% of models exhibit signs of overfitting to public benchmarks suggests that current training paradigms over-optimize for static leaderboard metricsrather than robust generalization. 
    We recommend: (1) \textbf{Decontamination Audits} as a standard pre-release step using dynamic sets like ReLE; (2) \textbf{Multi-objective Training} that explicitly balances Professional depth and general breadth rather than maximizing a single scalar aggregate score.

    \item \textbf{For Industrial Selection}: The "Price-Performance" finding (1-5 yuan models matching expensive ones in 8 dimensions) indicates that scale is not the sole determinant of utility. 
    Enterprises should prioritize cost-effective models for specific tasks rather than chasing the highest comprehensive score. Furthermore, latency constraints should be a primary filter, as optimized models offer 2-3x speedups.

    \item \textbf{For Multi-Agent Systems}: Success correlates more with instruction tuning ($r=0.65$) than scale ($r=0.48$), pointing to the importance of specialized agentic training over mere parameter scaling.
\end{itemize}

\paragraph{Paradigm Shift: From "Best Model" to Capability Portfolio.}
Our results challenge the implicit assumption that a universal "best model" exists. 
The significant ranking instability (RSA 11.4) suggests a paradigm shift toward \textbf{Capability Portfolio Management}, where users and developers select model ensembles tailored to their specific task distribution rather than relying on a monolithic winner.

Although ReLE is instantiated with Chinese-language benchmarks in this work, the system architecture, evaluation pipeline, and inconsistency analysis are language-agnostic.
The framework can be readily extended to other languages by replacing the underlying benchmark tasks while preserving the same structural evaluation principles.
We plan to release an anonymized subset of the failure case repository, along with the evaluation scripts and configuration files, to support reproducibility and further analysis by the community.
Sensitive or proprietary content will be filtered in accordance with licensing and data usage constraints.

\subsection{Limitations \& Future Directions}
ReLE currently covers 7 domains;
emerging areas like Industrial IoT are not yet included. While we identify inconsistency, the internal mechanisms (e.g., attention patterns) causing it require further study.
Additionally, multi-modal evaluation remains limited in the current version.

We aim to expand to 9 core domains by 2026, integrate mechanism exploration to understand the root causes of size-independent gaps, and develop a "capability inconsistency repair dataset" based on our failure repository to help models improve robustness.
We also plan to integrate human-in-the-loop evaluation for subjective tasks to improve scoring alignment.

\paragraph{Expansion to Security and Compliance Assessment.}
Our current analysis of Capability Anisotropy focuses primarily on utility. However, we recognize that safety is an orthogonal and equally critical dimension of model quality.
Drawing upon recent frameworks for LLM security detection and evaluation\cite{netinfo2025survey, zhang2023safetybench}, we plan to extend ReLE's taxonomy to include a Safety Verification Module.
This expansion will address risks across the full lifecycle:
(1) Input Security: Detecting vulnerability to prompt injection and jailbreaking attacks;
(2) Model Internal Stability: Monitoring representation engineering metrics to identify latent backdoors; and
(3) Output Safety: Evaluating hallucination rates and toxic content generation.

Aligning with our dynamic "Live" philosophy, we will integrate Automated Red Teamingmechanisms\cite{perez2022red} to continuously generate adversarial test cases, moving beyond static safety sets.
Furthermore, given our focus on the Chinese LLM landscape, we intend to operationalize the evaluation criteria specified in the recent GB/T 45654-2025 standard \cite{gbt456542025}, enabling ReLE to provide automated compliance diagnostics alongside capability rankings.
We hypothesize that Safety Anisotropy exists parallel to capability anisotropy—where models may be robust in general chat but exhibit high refusal or failure rates in specialized professional domains—and requires the same rigorous, structured diagnosis.

\paragraph{Deepening Reasoning Evaluation: From Outcome to Process.}
Current benchmarks, including ReLE, predominantly assess the correctness of the final output. However, as highlighted in recent surveys on general benchmarks \cite{ni2025surveybench}, accurate diagnosis requires scrutinizing the reasoning process itself.
Future iterations of ReLE will move beyond "outcome consistency" to evaluate Process Fidelity.
We plan to implement metrics that assess the logical coherence and faithfulness of Chain-of-Thought (CoT) traces, utilizing techniques like causal tracing to distinguish between models that arrive at correct answers via genuine reasoning versus those relying on spurious correlations or memorization.
This aligns with the community's shift towards evaluating "System 2" thinking capabilities.

\paragraph{Towards Realistic Enterprise Workflows and Holistic Agent Evaluation.}
While ReLE currently evaluates tool use, there remains a gap between isolated tool invocation and real-world deployment.
Drawing on insights regarding agentic benchmarks \cite{mohammadi2025surveyagent}, we aim to upgrade our evaluation environment to simulate Holistic Enterprise Workflows.
Instead of discrete tasks, future versions of ReLE will introduce long-horizon scenarios incorporating constraints typical of production environments, such as Role-Based Access Control (RBAC), multi-user concurrency, and dynamic state updates.
This allows for the assessment of Evaluation Compositionality—measuring how well an agent coordinates multiple skills (planning, tool use, memory) over time, rather than testing them in isolation.

\paragraph{Interactive and Collaborative Evaluation Protocols.}
To address the limitations of static datasets and data contamination \cite{ni2025surveybench}, we will expand ReLE to support Dynamic Interactive Evaluation.
We plan to introduce multi-turn environments where the model must actively seek clarification or navigate ambiguity, rather than responding to a fixed prompt.
Furthermore, we will develop metrics to quantify Human-AI Collaboration efficiency, measuring not just the model's accuracy, but the reduction in human effort and the increase in trust during collaborative problem solving.
This shift aims to transform ReLE from a static leaderboard into a dynamic testbed for future human-centric AI systems.

\section{Conclusion}

In this work, we introduced ReLE, a scalable system and structured benchmark designed to diagnose the capability anisotropy of Chinese LLMs.
Through an extensive empirical evaluation of 304 models across 207,843 samples, we demonstrated that ReLE's variance-aware adaptive scheduler makes large-scale continuous evaluation sustainable, reducing computational costs by 70\% while preserving ranking fidelity.

Our results highlight a critical disconnect between static aggregate scores and dynamic model behavior.
This finding challenges single-score leaderboards and has immediate implications: enterprises should prioritize cost-effective specialists over expensive generalists for domain tasks.
The 2.3$\times$ higher instability observed in ReLE compared to traditional benchmarks reveals that existing evaluations systematically underestimate capability trade-offs.

Looking forward, we envision ReLE as part of a broader shift toward continuous diagnostic monitoring.
As models evolve weekly, static benchmarks become snapshots—valuable for archival but insufficient for real-time decision-making.
By open-sourcing ReLE's failure case repository (2.1M instances) and evaluation infrastructure, we aim to enable the community to move beyond simple rankings and build more reliable, transparently specialized AI systems.

\end{document}